# A Review of Machine Learning Techniques in Imbalanced Data and Future Trends

Elaheh Jafarigol[a,1], Theodore B. Trafalis[a], Neshat Mohammadi[a]

[a]*School of Industrial and Systems Engineering University of Oklahoma, 202 W. Boyd St., Room 124, Norman, Oklahoma 73019, USA*

---

**Abstract**

For over two decades, detecting rare events has been a challenging task among researchers in the data mining and machine learning domain. Real-life problems inspire researchers to navigate and further improve data processing and algorithmic approaches to achieve effective and computationally efficient methods for imbalanced learning. In this paper, we have collected and reviewed 258 peer-reviewed papers from archival journals and conference papers in an attempt to provide an in-depth review of various approaches in imbalanced learning from technical and application perspectives. This work aims to provide a structured review of methods used to address the problem of imbalanced data in various domains and create a general guideline for researchers in academia or industry who want to dive into the broad field of machine learning using large-scale imbalanced data.

---

## Introduction

Classification problems are a major part of supervised learning and very often, the data is not equally distributed between the classes. The performance of the classifier is affected by the ratio of the majority class to the minority class, hence misclassification is more severe when the data is extremely imbalanced [1, 2, 3, 4, 5, 6]. In addition to the relative proportion of classes, the absolute number of available instances in the minority class is also an important factor. The problem with imbalanced data is magnified when the minority class consists of rare events. Rare events are defined as events that occur significantly less often than common events. In the case of rare events, classification becomes more challenging, because the classifier is often overwhelmed by the majority class and the results are biased. Therefore, without a significant loss in overall accuracy, the minority class is misclassified. Based on the type of data, the size of the data set and the distribution of data between classes, the issue of imbalanced learning can appear at different levels. The problem definition issues are caused by a lack of adequate information about the minority class[7]. Problem definition issues can cause evaluation metrics such as accuracy and error rate to fail in representing the minority class. Therefore, other evaluation metrics are defined to measure the classifier in imbalanced learning problems.

[1] Corresponding author
*Email address:* elaheh.jafarigol@ou.edu (Elaheh Jafarigol)

The data issues are the result of absolute rarity and extremely imbalanced data. Resampling methods are the standard solution to this issue. Algorithm issues are caused by inadequacies of the learning algorithm and may result in poor classification accuracy of the minority class. Such issues are caused by the model's failure in learning the necessary criteria for classification. The goal of imbalanced learning is to find an optimal classifier that is capable of providing a balanced degree of predictive accuracy for the minority class as well as the majority class [8, 9, 10, 11, 12, 13, 14]. These methods are primarily attempting to address the issue of absolute class imbalance that exists in some datasets. However, the relative class imbalance is still an important issue in datasets where we have an abundance of training examples, but the distribution of the different classes might be severely skewed. In this latter situation, one can have access to enough examples from the minority class, even if the frequency of the minority class is very small, as long as the total number of examples is sufficiently big [15]. With the broad applications of imbalanced learning in the real world, this area has attracted the interest of many researchers and despite the advances, most imbalanced learning methods are still sensitive to highly imbalanced data. In this survey, we have selected 258 peer-reviewed papers among the papers published on the topic of imbalanced learning and its applications. Figure 1 presents the technical key words used in our search.

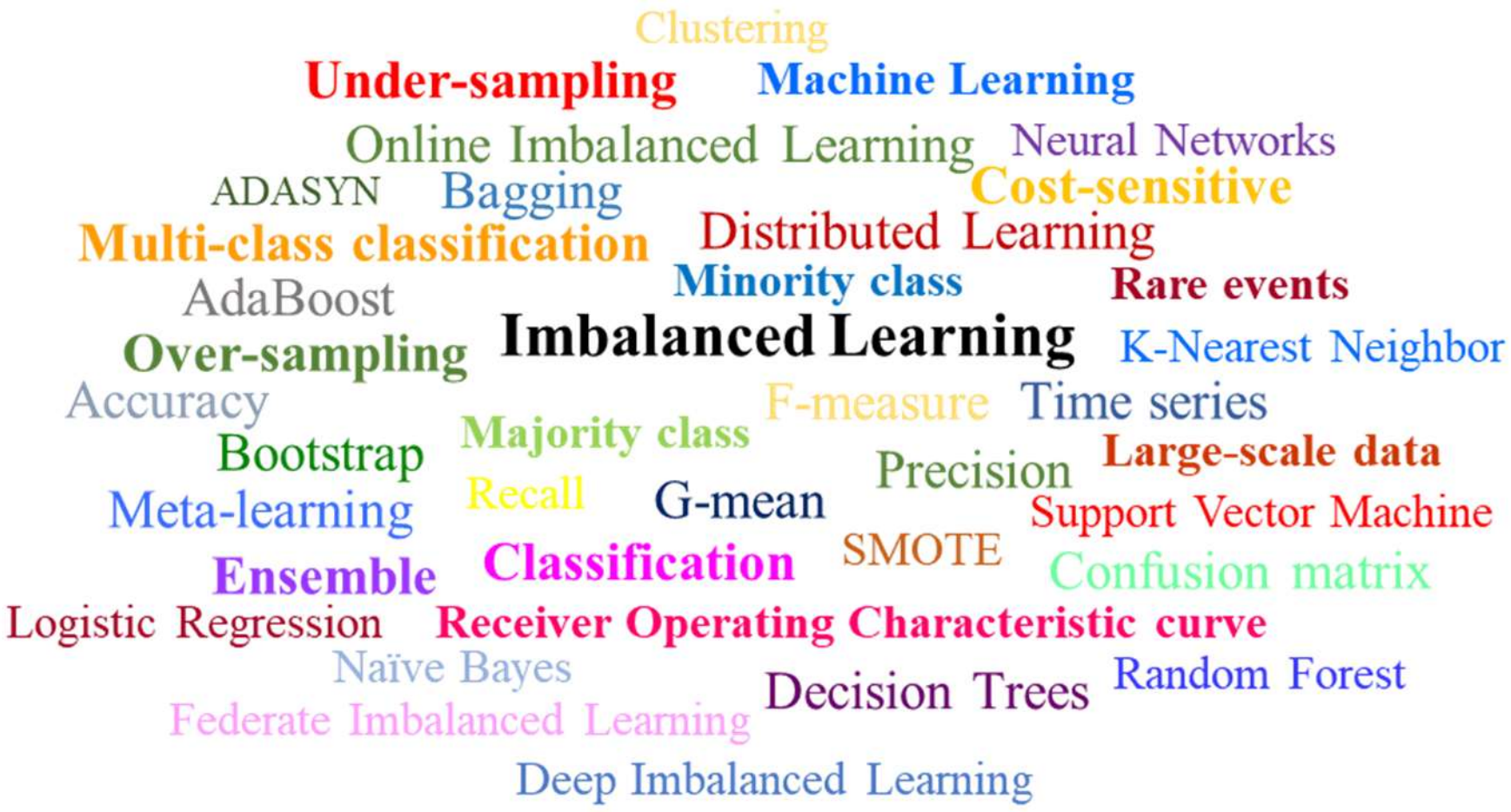

Figure 1: Technical Keywords in Imbalanced Learning Literature

In this paper, an overview of different approaches for the problem of imbalanced learning categorized based on the format provided in Figure 2, followed bu its applications in real-life problems is presented. The paper is organized as follows, in section 1, we provide a categorized definition of problem definition approaches and the different types of metrics used in this setting. In section 2, we focus on data processing approaches and extensively study different over-sampling and under-sampling methods used in the literature. Section 3 focuses on the algorithmic approach and the core machine learning methods for learning from large imbalanced datasets. In section 4 an overview of imbalanced learning applications is provided. Finally, we discuss some ideas of future research trends and conclude the paper.

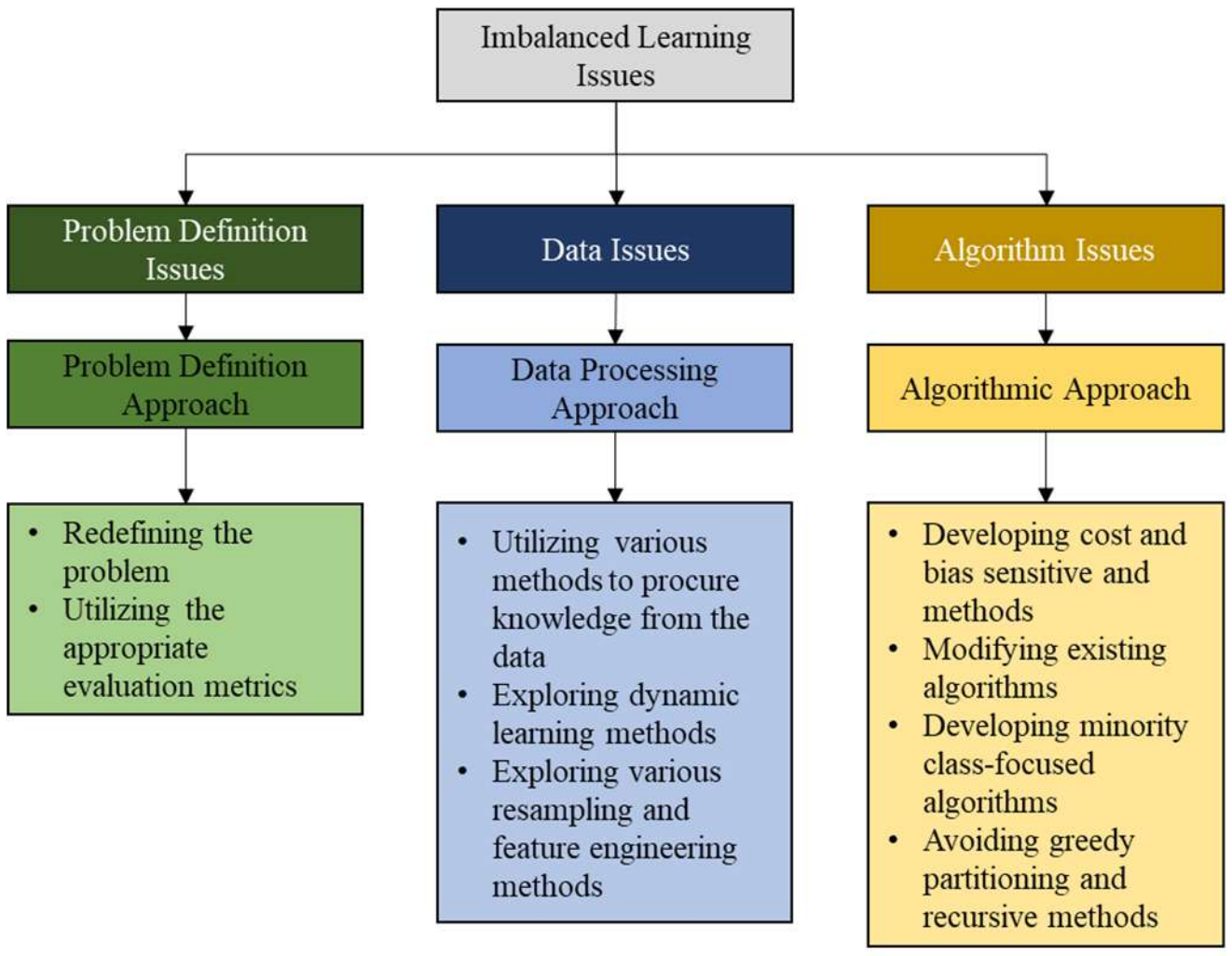

Figure 2: General Approaches in Imbalanced Learning

## Current Approaches

## 1. Problem Definition Approaches

### *1.1. Evaluation Metrics*

Evaluation is an important part of the learning process. Evaluation metrics are generally used to assess the generalization ability of the learning method on test data. One of the major issues that arise with imbalanced data is the inadequacy of well-known metrics such as accuracy and error rate in the evaluation of classification performance. Appropriate evaluation metrics are important in evaluating the quality of learning. Therefore, several authors have addressed this major issue and a new set of functions have been defined to determine how the classifier performs in classifying imbalanced data [16, 17, 18]. The authors of the paper published by Ferri et al. [19] have used experimental and theoretical analysis to compare and rank the evaluation metrics that work best on evaluating the learned model on imbalance data and analyze the identifiable clusters and relationships between the metrics. These experiments provide recommendations on the metrics that would be more appropriate for any specific application. Evaluation metrics are categorized into three types in the literature; threshold, probability, and ranking metrics [20].

### 1.1.1. *Threshold Evaluation Metrics*

The threshold type of evaluation metric is defined based on a confusion matrix as it is shown in Table 1. In binary classification, given that the predicted value of test samples in the majority class is denoted as N (Negative), and the predicted value of the test samples in the minority class is denoted as P (Positive), the Confusion matrix is defined. Note that, the definition of a Confusion matrix can be extended to multi-class classification as well.

Table 1: Confusion Matrix

| | | Predictive Value | |
|---|---|---|---|
| | | Positive | Negative |
| Actual Value | Positive | *TP* | *FN* |
| | Negative | *FP* | *TN* |

Based on this notation, Accuracy is defined as a measure for performance of a classification algorithm. Accuracy is defined as:

$$Accuracy = \frac{TP + TN}{TP + TN + FP + FN} \quad (1)$$

Accuracy is easy to use and interpret, however, despite being widely used by practitioners, it cannot provide enough information to ensure a reliable learning method when the data is imbalanced [21].

Classification performance metrics for imbalanced learning based on the confusion metrics are defined as:

$$Precision = \frac{TP}{TP + FP} \quad (2)$$

and,

$$Recall = \frac{TP}{TP + FN} \quad (3)$$

Precision and recall have an inverse relationship and when used together can provide valid insight into the performance of the classifier with regard to the minority class. Precision and recall measure how exact and complete the model is, respectively. Also, the precision-recall curve allows us to study the changes in both metrics simultaneously. In imbalanced learning, models with high recall on the minority class and high precision on the majority class are desired. Thus, F-measure is a valuable evaluation metric in imbalanced learning defined as:

$$F - measure = \frac{(1+\beta)^2 Recall * Precision}{\beta^2 Recall + Precision} \quad (4)$$

where $\beta$ is the relative importance of precision versus recall, and it is usually set equal to one. In the presence of rare events a common approach is to maximize the F-measure. Musicant et al. [22] have developed an approach to maximize the F-measure by using SVMs. Geometric mean/ G-mean is an important evaluation metric that is used explicitly for imbalanced learning.

$$G - mean = \sqrt{\frac{TP}{TP + FN} \cdot \frac{TN}{TN + FP}} \quad (5)$$

A high G-mean indicates that the model is performing well in both classes. Other metrics used in the literature are but not limited to:

$$Sensitivity = \frac{TP}{TP+FN} \quad (6)$$

$$Specificity = \frac{TN}{TN+FP} \quad (7)$$

$$NegativePredictiveValue = \frac{TN}{TN + FN} \quad (8)$$

Mathews Correlation Coefficient (MCC) [23] is defines as:

$$MCC = \frac{TP * TN - FP * FN}{\sqrt{(TP + FP)(TP + FN)(TP + FN)(TN + FN)}} \quad (9)$$

and Bookmaker Informedness/Youden's index = Sensitivity + Specificity – 1 [24].

#### 1.1.2. *Probability Evaluation Metrics*

The probability evaluation metrics are used with classification problems that focus on predicting the probability of a class label. Two popular probability predicting models are regression models and Artificial Neural Networks (ANN) [25]. Minimum Risk Metric (MRM) utilizes the posterior probability estimations to minimize the misclassification risk and provide an optimal solution. There are several probabilistic evaluation metrics such as Short and Fukunaga Metric, the Value Difference Metric, and Euclidean-Hamming metrics. The Short and Fukunaga, the Value Difference and Euclidean-Hamming metrics are distance functions used in Nearest Neighbor (NN) learning models to measure the distance between two instances, that can determine the associated attribute and classify the instance in the test data [26]. Log Loss is a classification performance metric based on the cross-entropy function. Given that the expected/known probability of an instance in the training data is denoted as P and the predicted probability of an instance in the test data is denoted as Q, the cross-entropy for an instance in binary-classification is defined as:

$$H(P,Q) = -(P(class0) * log(Q(class0)) + P(class1) * log(Q(class1)) \quad (10)$$

In this equation, the probability P is defined based on the Bernoulli distribution for the positive class and natural logarithm. When the instance is known, the cross-entropy is zero, therefore, we try to minimize the cross-entropy of the model.

### *1.1.3. Ranking Evaluation Metrics*

Sensitivity or TP rate and Specificity or TN rate are used to define the Receiver Operating Characteristic (ROC) curve, which is a visual representation of the classification performance. The Area Under the Curve (AUC) is defined as $\frac{Sensitivity+Specificity}{2}$. AUC does not depend on the classifier and it is a reliable tool for model comparison because it is scale-invariant, and the output is the ranking of classifiers rather than their absolute value. AUC can also assess the quality of models using a threshold-invariant [27, 28].

Although AUC is widely used for the evaluation and discrimination process of binary classification models, it can be misleading sometimes. AUC uses a different misclassification cost for each classifier. Some researchers have addressed this issue and proposed modifications or alternative metrics such as the H measure which uses a symmetric Beta distribution in the AUC [29, 30].

To summarize, we have presented the evaluation metrics categorized based on their outcome in imbalanced learning studies in Tables 2 ,3, 4 and Figure 3.

Table 2: Threshold Metrics in Supervised Learning

| Metrics | Definition |
|---|---|
| Accuracy | The ratio of the correctly classified instances over the total number of classified instances |
| Error Rate | The ratio of misclassification errors over the classified instances |
| Precision | The proportion of instances that were labeled correctly among those with the positive label in the test data |
| Recall | The portion of positive instances in the test data that were labeled correctly |
| F-measure | The trade-off between precision and recall |
| G-mean | The measure to maximize the accuracy of the model over each class by considering both classes for evaluation |
| Sensitivity | The relative performance of the classifier over the minority class |
| Specificity | The relative performance of the classifier over the minority class |
| Negative Predictive Value | The number of TN over the instances with positive label in the test data |
| Mathews Correlation Coefficient | The measure of quality in binary classification |
| Bookmaker Informedness | The measure if discrimination capability of the classifier |

Table 3: Probabilistic Metrics in Supervised Learning

| Metrics | Definition |
|---|---|
| Minimum Risk | The probability of minimizing the misclassofocation risk while maintaining an optimal solution |
| Short and Fukunaga | A measure of distance between instances in Nearest Neighbor models |
| Euclidean-Hamming | A measure of distance between instances in Nearest Neighbor models |
| Log-loss | The negative log-likelihood under the Bernoulli distribution |

·: Ranking Metrics in Supervised Learning

| Metrics | Definition |
|---|---|
| ROC Curve | Evaluate and rank several classifiers |
| AUC | The probability of correctly classifying the positive instances while the number of false positives is minimized |

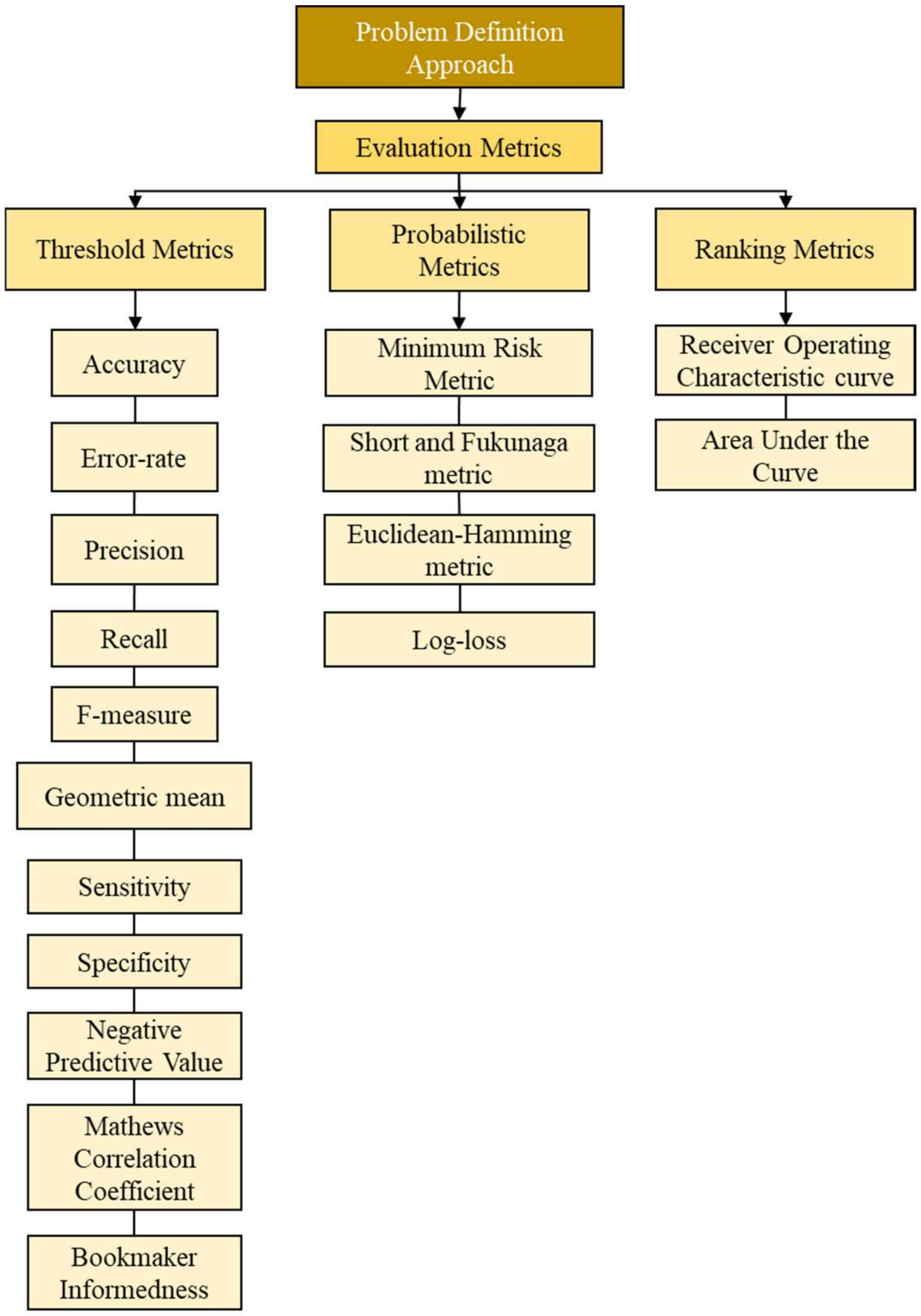

Figure 3: Evaluation Metrics in Imbalanced Learning

## 2. Data Processing Approaches

### *2.1. Resampling Methods*

Resampling methods are developed to balance the ratio of the classes in imbalanced learning by adjusting the minority class or the majority class and enhancing the performance of the classifier [31]. Generally, basic resampling methods follow two strategies. The first strategy is removing instances from the majority class known as Random Under-sampling (RUS) [32, 33]. The second is adding new instances to the minority class, known as Random Over-sampling (ROS)[34]. These methods can be utilized on their own or in combination with each other to adjust the distribution of the data before classification. A limitation of RUS and ROS is removing valuable information in the resampling process, therefore, under-fitting or over-fitting the data, respectively. To avoid such issues, advanced resampling methods were developed based on the idea of a guided resampling. Advanced resampling methods include multiple variations of under-sampling and over-sampling methods [35, 36]

#### *2.1.1. Under-sampling Methods*

Under-sampling following the Nearest Neighbor (NN0 rule is the classification of data based on the similarities between the data point and its nearest neighbor. This decision rule has a lower probability of error than several other decision rules. Variation of under-sampling based on NN rule includes condensed NN method, the edited NN method, the repeated edited NN method, and neighborhood cleaning method [37] and other variations [38, 39]. Tomek's links (T-link) is an enhancement of NN rule for under-sampling the majority class in which the pair of data with opposite labels in the same neighborhood create a Tomek link. The data point on the link that belongs to the majority class is removed. This method improves the classification accuracy of the minority class by creating a distinct margin between the two classes [40]. Under-sampling based on clustering utilizes the clustering algorithms such as K-means that show promising performance with imbalanced data [41]. The one-sided selection method is an adaptation of Tomek's link. In this method, a subset of the majority class is selected for classification while the minority class remains untouched [42]. Under-sampling based on Instances Hardness Threshold (IHT) method is used to overcome the problem of imbalanced data. This under-sampling method reduces the size of the majority class by removing the data that has a high hardness threshold, which is the probability of misclassification of the data [43, 44, 45].

### 2.1.2. *Over-sampling Methods*

An effective way of dealing with the issue of imbalanced data is Over-sampling. Studies suggest that the number of features and imbalance ratio are important factors in determining the best approach [46, 47]. Over-sampling methods such as bootstrap-based over-sampling, over-sampling based on Synthetic Minority Over-sampling Technique (SMOTE), and over-sampling based on Adaptive Synthetic sampling method (ADASYN) are widely used in imbalanced learning [48, 49, 50, 51, 52, 53, 54] Bootstrap-based over-sampling is iteratively replicating the instances of a selected sample, in which the instances are replaced and are probable to be selected more than once. The number of iterations and the sample size is required before oversampling [55, 56].

In over-sampling using SMOTE, the number of instances in the minority class is increased by syntactically creating new instances instead of merely replicating the existing instances. SMOTE generates data in the feature space, and it depends on introducing new instances based on the nearest neighbors [57]. In this method, the new examples are added near the line segment that joins the nearest neighbors of the minority class. The nearest neighbors are selected to create the instances required for over-sampling [58, 59, 60, 61, 62, 63, 64, 65, 66]. Inspired by SMOTE, XiChen et al. [67] proposed a sampling method, in which new synthetic neighborhood samples are generated. Controlling the number of generated samples can improve the balance ratio and promote diversity in the data. Zhou et al. [68] proposed a cost-sensitive SMOTE for data classification. Since the samples are generated in the feature space, creating a new sample in a nonlinear space can improve the results after resampling of the minority class [69].

Among different variations of SMOTE, ADASYN, motivated by SMOTE, is a popular oversampling method, in which the data is synthetically generated to increase the size of the minority class. The size of the generated data is determined by the density distribution criteria defined for each example which is an advantage over SMOTE in that the number of generated data is predetermined [70, 71, 72]. The over-sampling methods are not limited to the ones mentioned in this paper. These methods can be applied alone or in combination with each other to improve classification results [73]. For example, in this paper, the authors combined the ADASYN method with a cost-sensitive base model to improve the results in a study of the transient stability of power systems [74].

Many studies have been carried out to evaluate the effectiveness and efficiency of resampling methods, and provide a guideline on selecting the best one for the specific data [75, 76, 77, 78]. Figure 4 provides a structured overview of resampling methods used in data processing approach. Over the years, different variations of resampling methods are used in combination with algorithmic approaches to enhance the prediction accuracy in imbalanced data .

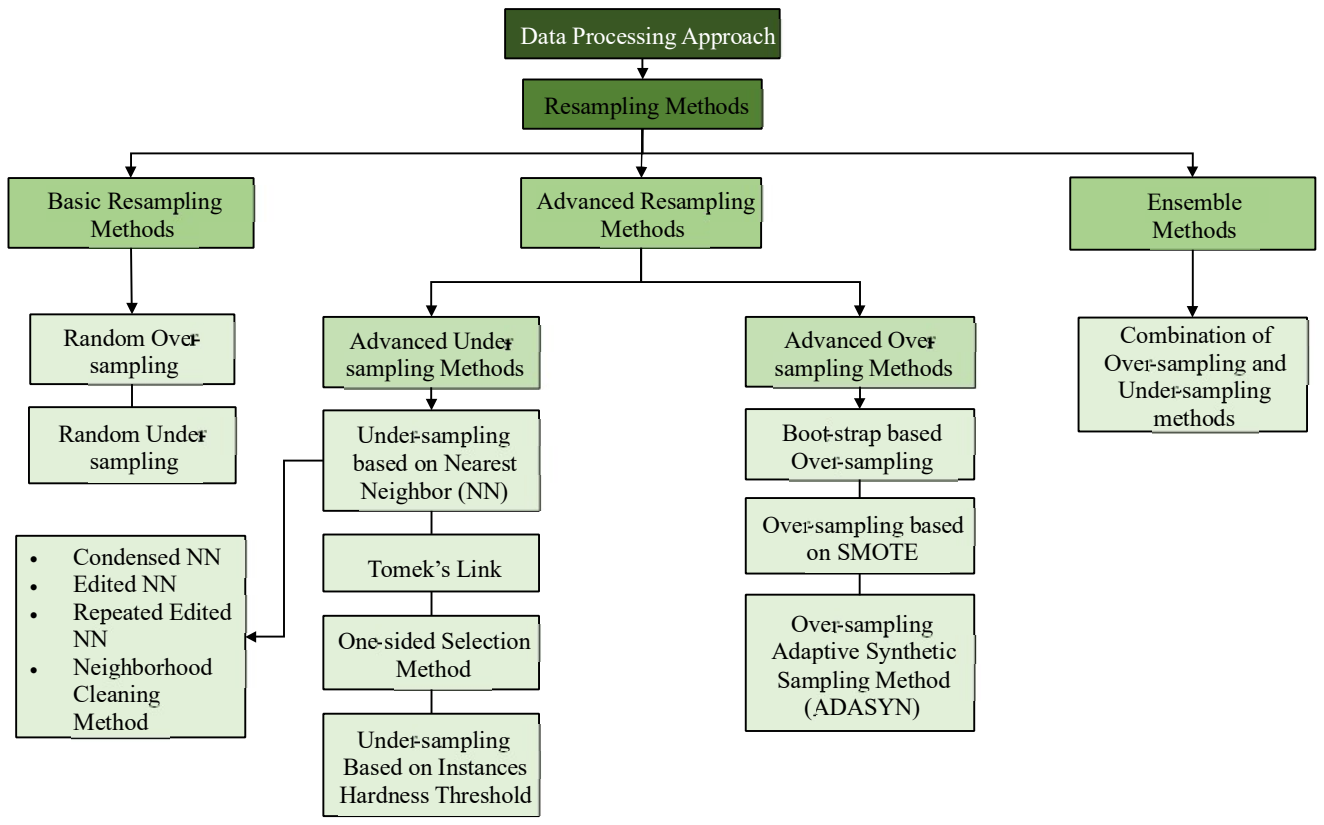

Figure 4: Data Processing Methods in Imbalanced Learning

## 3. Algorithmic Approaches

### *3.1. Cost-sensitive Methods*

In real-world applications of imbalanced learning, such as cancer diagnosis, fraud detection, or severe weather prediction; the misclassification cost is different for the minority class and the majority class, respectively. In imbalanced learning where the misclassification cost of the minority class is more important, cost-sensitive methods are used. In cost-sensitive methods, the cost of misclassification is known and defined in a cost matrix based on the cost associated with a false positive and a false negative. The goal is to classify the data while minimizing the expected misclassification cost of making a false prediction. In imbalanced learning, we can benefit from shifting the classification algorithm towards further minimizing the misclassification error of the minority class [79][80]. Cost-sensitive methods are categorized as direct and meta-learning methods.

### *3.1.1. Direct Methods*

In the direct methods, the classifiers are designed to anticipate different misclassification costs for false positives and false negatives. Cost-sensitive decision trees are an example of such methods that have improved the classification results by incorporating the cost in the model and aiming to minimize the misclassification cost [81, 82, 83, 84, 85, 86]. In other Cost-sensitive methods, weights are used in the classification algorithm [87]. Studies show that iterative weighting of the samples can improve the results as well as achieving computational efficiency [88, 89]. Cost-sensitive boosting methods have also been used to compare the effectiveness of such algorithms on benchmark datasets [90, 91]. Wu et al. [92] used a cost-sensitive multi-set feature learning on multiple samples constructed by partitioning the majority class and combining the blocks with the minority class to obtain balanced datasets. The model is evaluated using benchmark data sets and recommended for highly imbalanced data. One of the challenges of cost-sensitive methods is identifying the misclassification costs.

Zhang et al [93] proposed an adaptive differential evolution to find the optimal misclassification costs.

### *3.1.2. Meta-learning Methods*

Meta-learning methods are used to convert cost-insensitive classifiers into cost-sensitive algorithms without making modifications to the algorithm by thresholding and sampling methods. Thresholding models classify the data by producing probability estimations using a cost-insensitive algorithm and use a threshold to classify the data [94, 95, 96]. Thresholding is an effective method that expands the space and increases the probability of classifying the instances associated with the minority class. Therefore, they often produce the lower misclassification cost comparing to other classification methods [97]. Sampling meta-learning methods modify the class distributions in the dataset, before training the data using a cost-insensitive classifier. Weighing is also a type of sampling method in which a normalized weight based on the misclassification cost is assigned to the data before classification[98, 99]. Cost-sensitive learning is most effective when embedded in the machine learning based model, categorized as Ensemble methods. Ensemble methods are discussed thoroughly in a separate section.

### *3.2. Machine Learning Based Modeling*

Various machine learning models have been explored in the attempt to minimize the misclassification error of the minority class such as Logistic Regression (LR), Artificial Neural Networks (ANNs) [100], Random Forest (RF), Decision Trees (DT), Naive Bayes (NB), and Gaussian NB, and K-Nearest-Neighbor (KNN), Support Vector Machines (SVM).Empirical studies on benchmark data suggest different base predictor models [101, 102].

#### *3.2.1. Tree-based Models*

DT is a classification algorithm that splits the data set into smaller subsets to predict the output value of the test data. The conditions by which the data is split are called leaves, and the decision is known as a branch. The data is split until we have reached the depth of the tree and no further split is possible. DT is a fast and simple algorithm in which the process of classification and inquiries made are clear [103, 104, 105, 106, 107, 108].

RF is a powerful ensemble method, which is an aggregation of less accurate predictive models to create a better model. This model is used for regression or classification. In RF classification, decision trees are used to introduce randomness when selecting the suboptimal splits, and the goal is to aggregate as many uncorrelated trees as possible and improve the accuracy at each step [109, 110, 111, 112, 113].

#### *3.2.2. Probabilistic Models*

NB is a supervised learning method. The first assumption in this method is that all the data points are independent of one another. This is an unrealistic yet helpful assumption for training the data. In this method, the training data is used to calculate the probability of each class and the conditional probability of each class for a given data point. These two pieces of information are used to predict the class of new data points. Gaussian NB is a modification of the NB method, except that for the input data in real values, a Gaussian distribution is assumed to make calculating the probabilities easier [114, 115]

### *3.2.3. Neighborhood-based Models*

KNN is a simple yet powerful algorithm that uses the whole dataset for classification. To classify a new data point, KNN uses the data points closer to the designated point based on their Euclidean distance. Then it summarizes their output values and assigns the result as the label of the new data point. In KNN, training, and testing are combined in one step which increases the effectiveness and efficiency of the model which is one of the widely used imbalanced learning models. The papers [116, 117, 118, 119, 120, 121, 122, 123, 124, 125, 126, 127] are representative of those methods.

### *3.2.4. Kernel-based Models*

Linear and logistic regression are probably the most well-known and widely used machine learning algorithms. Linear regression is used for predicting values that are in a range, but logistic regression is appropriate when we are trying to predict categorical output values such as binary classification [128]. LR is presented by a non-linear function, and the data is classified based on the features correlated with the output variable [129, 130, 131]. To further improve the traditional LR, Ohsaki et al. [132] proposed a novel confusion-based kernel logistic regression that utilized a harmonic mean objective function to improve generalization and classification errors of the model. Historically, in the 1950s and 1960s, perceptron algorithms were used for detecting linear relations in the data. Perceptron algorithm is one of the oldest machine learning algorithms. In this method, we need to associate a weight to the data points and define a threshold, known as bias. The weights and the threshold are extracted from the data. The weighted sum of the input data is calculated for predicting the output value. The label is one if the sum is greater than the designated threshold, and zero otherwise.

In perceptron algorithm, the goal is to find the set of weights that best classifies the data. Nieminen et al. [133] demonstrated the use of a single layer perceptron based on a multi-criteria optimized MLP as the base model. Although perceptron algorithms were useful for processing linear relations in the data, developing efficient and stable algorithms for detecting nonlinear relations was a major challenge for researchers at the time. In the mid-1980s, back-propagation Neural Networks (NNs) and decision trees revolutionized the field of non-linear pattern analysis. In the mid-1990s, kernel-based methods were developed for nonlinear data analysis while retaining the efficiency and stability of previous linear algorithms. Kernel-based methods apply to a broad range of data types such as sequence, text, image, graph, and vectors. They can detect different types of linear and non-linear relations and they are used for correlation, factor, cluster, and discriminant analysis. Kernel methods have a modular framework, in which first the data is processed into a kernel matrix, then the data is analyzed using various pattern analysis algorithms based on the information contained in the kernel matrix [134, 135, 136, 137, 138, 139]. Kernel matrix is obtained from mapping the data from the input space into a higher dimensional feature space, using a transformation function denoted as $\varphi(x)$. One of the challenges in kernel method is finding the kernel map, which is computationally expensive and sometimes impossible, therefore, kernel functions are defined by the dot product of the points in the input space. Using this feature, known as the kernel trick, $K(x_i,y_j) = \varphi(x_i)^T.\varphi(x_j)$, the data is mapped into the feature space, without explicitly defining the map. Different kernel functions have been developed. Kernel methods utilize a higher dimensional feature space to facilitate accurately classifying the minority class [140].

Among different classifiers, kernel-based SVM as introduced by Vapnik [141][142] has been widely used for imbalanced learning. SVM is a family of algorithms that use kernel methods to solve problems in classification and regression [143, 144, 145, 146, 147]. The idea of kernel SVM is to map the data to a higher dimensional feature space using a linear or nonlinear kernel [148, 149]. Then finding a separating hyperplane to maximize the margin of separation while minimizing the misclassification error by solving a quadratic optimization problem. SVMs are commonly used for classifying large data sets. The data is classified based on its location on either side of a hyperplane, which splits the input space. The separating hyperplane is not unique; however, the best hyperplane is the one that maximizes the margin of separation while minimizing the misclassification error. Combining cost-sensitive methods with SVM is a useful method for improving the misclassification cost [150, 151]. Cost-sensitive SVM embedded into the objective function can directly improve the classification performance when the feature set and tuning parameters are optimized [152]. Focusing on ROC and the AUC, Hu et al. [153] proposed the kernel online imbalanced learning algorithm that aims to maximize the AUC score while maintaining the regularization capabilities of the classifier. Weighted under-sampling SVM has improved the classification performance of SVM for imbalanced data [154]. Variations of SVM have been used for many applications such as fraud detection, gene profiling, weather prediction, etc [155, 156].

#### *3.2.5. Deep Imbalanced Learning*

ANN is a machine learning algorithm developed for separating non-linear data. In this method, a large number of units known as neurons are connected to form a multi-layer neural network. The neurons are divided into three types, input units that receive the information for processing, output units that contain the processing results, and the units in between known as hidden units. The efficiency of ANNs depends on the input units and their corresponding activation functions, the network architecture, and the weights of input connections, and the calculated weights of hidden units updated throughout the learning process.

ANNs apply to various real-world problems as well as imbalanced data [157, 158]. ANNs are studied with cost-sensitive methods to improve misclassification cost by moving the classification threshold closer to the majority class, which allows more instances to be classified as the minority class. Other methods are imposing greater weights on the samples associated with the minority class [159, 160]. Ya-Guan et al. [161] proposed an improved ANNs method called the Equilibrium Mini-batch Stochastic Gradient Descent that improves the model's training convergence error.

In recent years, Extreme Learning Machine (ELM) for NN structures proposed by Huang et al. [162] has been applied extensively for real-world imbalanced learning problems. Some of the proposed strategies based on ELM are Weighted ELM [163, 164, 165]. Class-specific ELM [166], and Class-specific Kernel ELM [167, 168] that have had promising results. When dealing with imbalanced data, deep learning algorithms face the same issues as traditional machine learning algorithms, and they fail to perform equally well in both classes [169]. Deep imbalanced learning models are developed to address the issue of imbalanced data in image recognition and computer vision [170**?** , 171].

Lin et al. [172] proposed a hybrid sampling method to remove the between class data points and guide the network to improve the classification results [173]. Dong et al.[174] developed a deep learning model to classify imbalanced datasets by imposing a class rectification loss as a regularization parameter to discover the boundaries in the minority class and reduce the effect of the majority class on the model. Bao et al. [175] introduced a deep learning framework to balance the data in a deeply transformed latent space. The superiority of this model is that feature learning, balancing, and discriminative learning are conducted simultaneously and it performs effectively on multi-classification problems. A cost-sensitive deep NN proposed by Khan et al. [176] is a robust feature representation of both classes. Therefore, it can have improved predictive capability. Lin et al. [177] proposed a deep reinforcement learning model based on the reward function specified for the minority and the majority class.

#### *3.2.6. Ensemble Methods*

Ensemble methods is an approach in machine learning that utilizes multiple machine learning based models to improve predictive accuracy. A group of ensemble methods is formed based on Bootstrap Aggregating (Bagging) in which several bootstrapped subsamples are created and trained using a base model [178, 179]. Later the model aggregates the decision tree models to create the optimal predicting ensemble method [180]. Random forest models are another type of ensemble method which is a variation of Bagging, in which splitting the tree based on different features creates a more accurate model. For imbalanced learning, cost-sensitive decision trees are introduced [181]. Another ensemble method called CSRoulette is introduced that improves the performance by producing samples of different sizes based on a cost-sensitive model, combined with Bagging [182]. An empirical study of ensemble methods and meta-learning methods suggests that although these methods are effective for binary classification of imbalanced data, they might not perform well on multi-class classification problems. For multi-classification problems, a combination of LR and KNN is used. In the LR part of the model, an ensemble of Bagging and Boosting methods has resulted in a promising outcome [183]. Variations of the Boosting algorithms such as Adaptive Boosting (Adaboost) have shown promising results in classifying imbalanced data [184]. A comparison of various ensemble methods suggests that combining data preprocessing approaches, such as RUS with Bagging or Boosting methods can result in higher performance [185]. An ensemble of random subsampling with RF has reasonable performance [186, 187]. Modifying the objective function to anticipate different factors to minimize the misclassification error in combination with evolutionary under-sampling methods is an example of ensemble methods aiming to improve the results of imbalanced learning [188]. Ensemble methods with multi-objective optimization function provide powerful algorithms for imbalanced learning [189, 190]. Ensemble methods are also effective for highly imbalanced data [191]. For image recognition, ensemble deep imbalanced learning with a focus on resampling the data, and a weighted loss function has improved the image classification results [192]. Wu et al. [193] used a genetic algorithm approach with a deep imbalanced learning model to optimize oversampling the minority class. Despite being able to improve the classification results, ensemble methods are computationally complex. A structured overview of the methods used in algorithmic approach is presented in Figure 5.

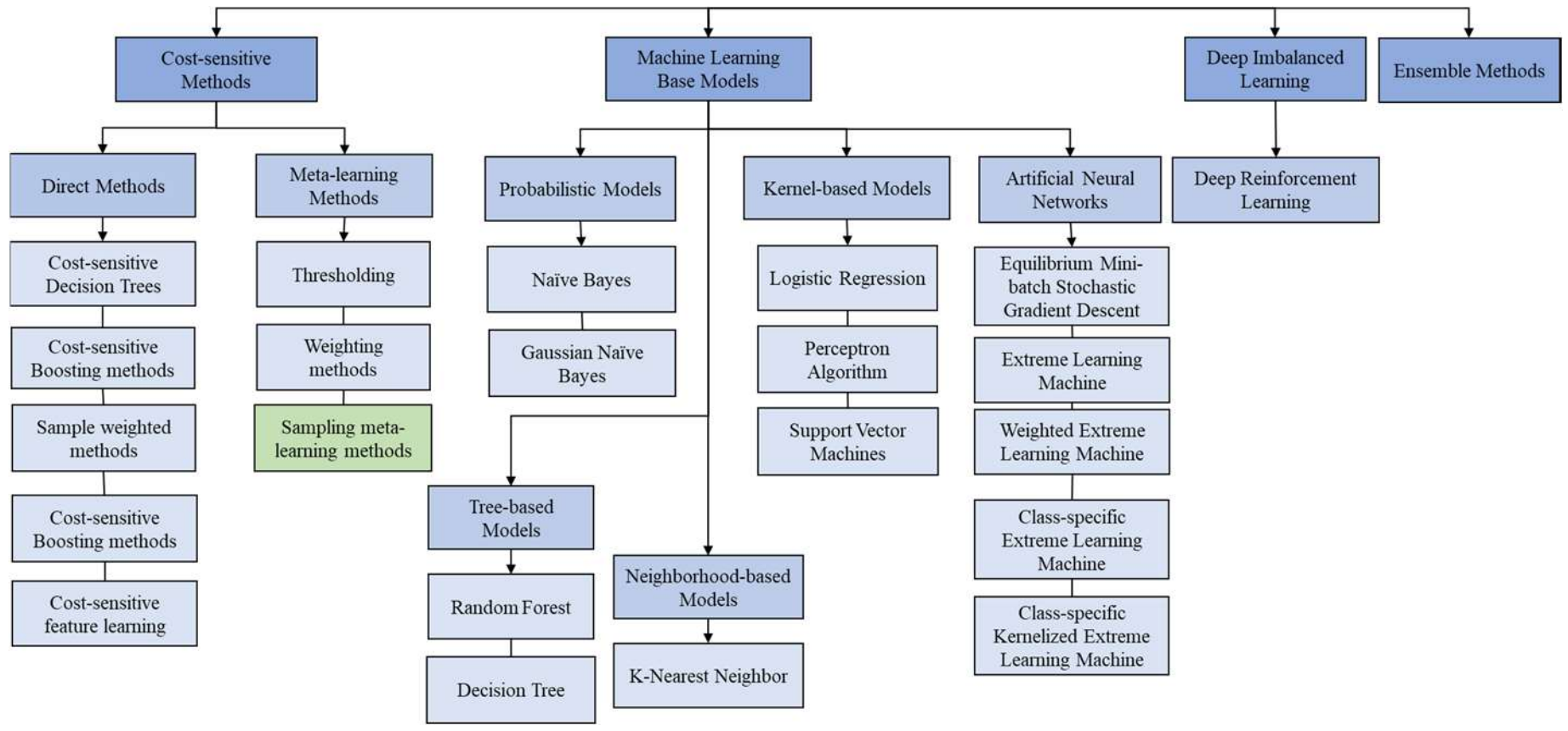

Figure 5: Algorithmic Approaches in Imbalanced Learning

## 4. Applications

Classification of imbalanced data is a challenging task and it is one of the popular research problems with many applications in the real world [194, 195]. In this section we have presented some of the highly impactful imbalanced learning problems, however, the applications of imbalanced learning are not limited to the mentioned examples. Figure 6 provides an overview of areas in which imbalanced learning is used.

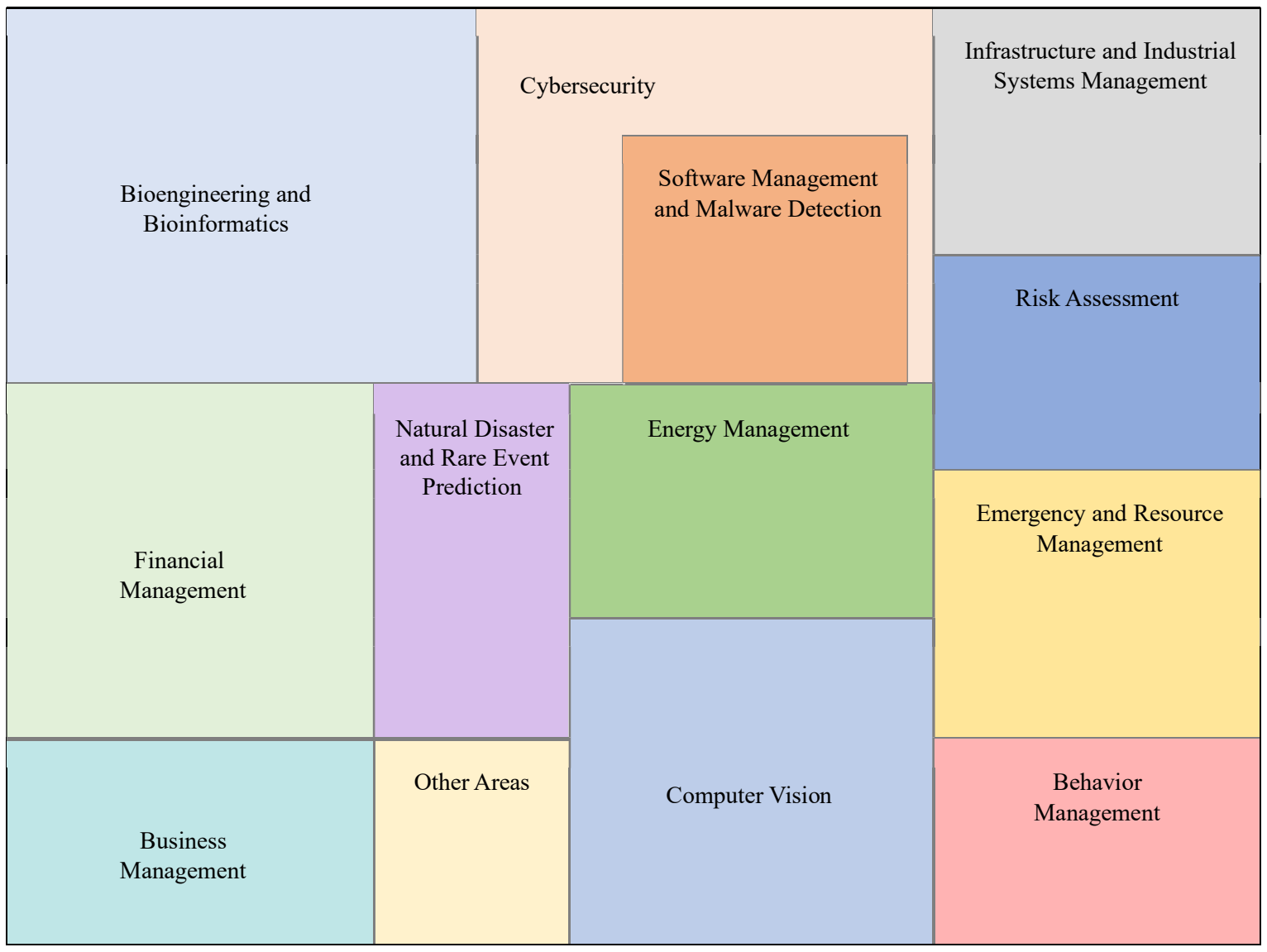

Figure 6: Applications of Imbalanced Learning

### 4.1. *Risk Assessment in Business and Finance*

In business analytics, bankruptcy prediction is an imbalanced learning problem. Gnip et al. [196] used multiple ensemble methods to accurately predict the data collected from medium-sized enterprises in the Slovak Republic [197]. In banking, credit scoring and evaluating the potential risk posed by applicants' unpaid loans is an important issue, and due to frequency, it is an example of imbalanced data [198, 199, 200]. For example, detecting fraudulent transactions using ensemble methods [201] and evaluating loan and credit applications can benefit from imbalanced learning to support the decision-making process. Approval or rejection of loans based on the applicant's credit history is an imbalanced learning problem with unpaid loan creating the minority class [202, 203, 204].

Fraud detection is one of the major applications of imbalanced learning algorithms. Bauder et al. [205] compared the performance of different resampling approaches on highly imbalanced data from Medicare to detect fraudulent cases.

### 4.2. *Behavior Management*

Imbalanced learning is also applicable to the data collected from Socioeconomic systems. For example, Orooji et al. [206] predicted the rate of high school dropout in Louisiana, US., which has negative impacts on the well-being of society, and Zheng et al. [207] explored a short tree-based adaptive classification test to assess the risk factors for juvenile delinquency.

### 4.3. *Cybersecurity and Software Management*

In cybersecurity, spam and software defect detection is an example of imbalanced learning [208, 209, 210, 211]. Chen et al. [212] proposed an ensemble model based on Choquet fuzzy integral with an improved SMOTE resampling technique for bug report identification that can prevent damage to software. Developing effective Intrusion Detection systems (IDS) is essential to cybersecurity [213]. Karatas et al. [214] used the SMOTE resampling method to improve IDS performance. Feng et al. [215] tackled the issue of imbalanced data in IDS classification using a cost-sensitive feature engineering method based on General Vector Machine(GVM) and Binary Ant Lion Optimizer. Zheng et al. [216] used a modified SVM to improve offline signature verification systems. Pang et al. [217] used an ensemble of SMOTE and SVM to detect malicious apps for android users.

### 4.4. Natural Disasters and Emergency Management

An impactful application of imbalanced learning is predicting rare natural disasters. Fernandez-Gomez et al. [218] studied the use of ensemble methods on predicting rare large magnitude earthquakes with a horizon of prediction of five days in Chile. Seismic capability evaluation of buildings is also an imbalanced learning problem in earthquake engineering [219]. Predicting severe weather events such as tornado is an imbalanced learning problem in meteorology and data mining [220, 221, 222]. Optimizing the available resources in urgent care is important in times of crisis. An ensemble method consisting of Bagging and DT can improve the prediction results for patient readmission to the emergency department of a hospital in Chile [223].

### 4.5. Bio-informatics and Bio-engineering

Medical diagnosis is an example of imbalanced learning in the field of bioinformatics and bioengineering. Zhang et al. [224], explored the use of an ensemble method of RUS with K-means and SVM to improve the diagnosis accuracy. Zheng et al. [225] used a Convolutional Neural Network (CNN) to detect exudate in optic images, that if detected correctly can prevent diabetic retinopathy and blindness. Jeong et al. [226] addressed the issue of multi-classification of imbalanced kidney data. In this paper, the glomerular rate is defined as target to diagnose chronic kidney disease. The goal is to classify the data into five stages using four methods of multinomial LR, and ordinal LR, RF, and Autoencoder (AE). The comparison of the four models suggests that AE provides better performance and is recommended for similar problems. Farhadi et al. [227] used a deep transfer learning model on constructing medical image data to evaluate the model's efficiency in diagnosing high-grade breast cancer. A breast cancer diagnosis has been improved by advanced imbalanced learning methods introduced in the past decade [228, 229, 230]. Other cost-sensitive methods have also been used to improve the classification accuracy of medical diagnosis [231, 232]. Deng et al. [233] have introduced a dynamic clustering method that iteratively adjusts the cluster based on the weight changes in the cluster. This algorithm is evaluated using gene expression cancer diagnosis data and applies to biological and cyber-physical systems. A deep imbalanced learning framework applies to different fields such as active balancing in biomedical data
[234].

*4.6. Computer Vision*

Image processing and recognizing facial images and other attributes in detail is a challenging task in computer vision, and the difficulties escalate when the data is imbalanced [235, 236, 237]. Various ensemble methods have been explored to classify multimedia data [238]. Pouyanfar et al. [239] proposed an ensemble deep learning framework based on the performance of SVM classifiers on deep feature sets which is evaluated using multi-media data for semantic event detection. In terms of application, different packages exist that can be used to implement the models in Python, R, or other scripting languages [240].

**Future Research Trends**

Learning from imbalanced data is one of the challenging tasks in data mining. However, it gets even more difficult when it is combined with other issues. Different studies have been carried out to explore strategies for specific issues of the minority class, such as highly imbalanced cases, noisy data [4], outliers, sparse data [241, 242] and the problem of imbalanced distribution within the minority class [243, 244]. Another category of imbalanced learning problems is multi-class problems that require more advanced techniques to deal with imbalanced data [245, 246]. Some of the proposed strategies such as weighted extreme learning machines, weighted support vector machines [247], and sequential ensemble learning have been relatively effective in the case of highly imbalanced data [248, 249, 250]. However, these methods are computationally complex and further improvement is desired. A real-world application of imbalanced learning is time series analysis with imbalanced and skewed data. This is particularly challenging due to the high dimension of the data and underlying correlations within the data, and further exploration is desired [251, 252, 253]. Imbalanced learning is an often over-looked issue in Online Learning of streaming data [254, 255]. Different methods such as cost-sensitive methods have been explored in various studies evaluated based on the imbalanced learning metrics [256].However, extensive research is required to address the issues in online imbalanced learning of large scale data. The last but not least is the problem of imbalanced Learning in distributed framework [257, 258]. Decentralized data centers often cause skewed class distribution in different classes. Distributed learning has gained more attention in the past few years and tackling the issue of imbalanced data in such framework is essential.

## Conclusion

Extensive research has been carried out to improve and identify the best approaches for imbalanced data in different fields from cyber-security to business analytic and bio-informatics. In this paper, we have provided a review of the wide range of methods applied to imbalanced data from a technical perspective. Examples of real-world applications have also been reviewed. We have collected and reviewed the papers published in peer-reviewed journals from 2000 to 2020 to understand the trends and advances in learning from imbalanced data and provide insights for future research trends in this highly anticipated field.